\documentclass{article}
\usepackage{spconf,amsmath,graphicx,kotex,amssymb,url}
\usepackage{xcolor}
\usepackage{booktabs}
\usepackage{comment}
\newcommand{\degree}{$^{\circ}$}
\newcommand{\etal}{\textit{et al}. }
\newcommand{\ie}{\textit{i}.\textit{e}., }
\newcommand{\eg}{\textit{e}.\textit{g}., }
\title{Upright Adjustment with Graph Convolutional Networks}
\name{Raehyuk Jung$^{*}$ $^1$, Sungmin Cho$^{*}$ $^2$, and Junseok Kwon$^2$
\thanks{$*$ Authors contributed equally.}
}
\address{Graduate School of Culture Technology, KAIST, Daejeon, Korea$^1$
\\
School of Computer Science and Engineering, Chung-Ang University, Seoul, Korea$^2$}
\begin{document}
%\ninept
%
\maketitle
\begin{abstract}
We present a novel method for the upright adjustment of 360\degree~images. 
Our network consists of two modules, which are a convolutional neural network (CNN) and a graph convolutional network (GCN). 
The input 360\degree~images is processed with the CNN for visual feature extraction, and the extracted feature map is converted into a graph that finds a spherical representation of the input. 
We also introduce a novel loss function to address the issue of discrete probability distributions defined on the surface of a sphere. 
Experimental results demonstrate that our method outperforms fully connected-based methods.
\end{abstract}

%%%%%%%%%%%%%%%%%%%%%%%%%%%%%%%%%%%%%%%%%%%%%%%%%%%%%%%%%%%%%%%%%%%%%%%%%%%%%%%%%%%%%%%%%%%%%%%%%%%%%%%%%%%%%%%%%
\section{Introduction}
\label{sec:intro}
\begin{figure}[t]
        \centering
        \includegraphics[width=1.0\linewidth]{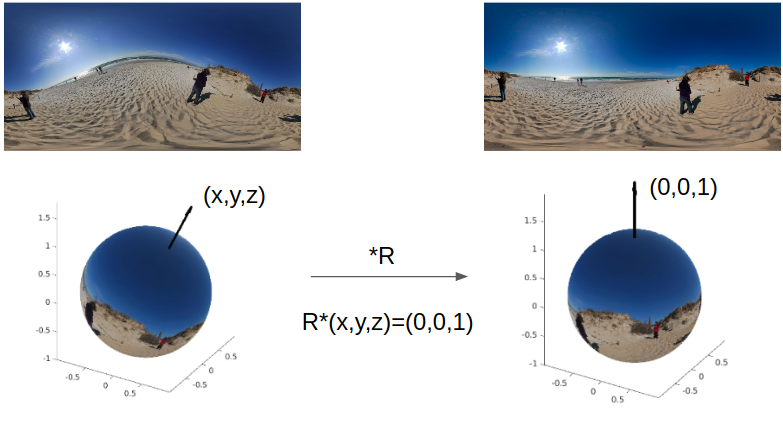}
    \vspace{-9mm}
    \caption{\textbf{Upright adjustment consisting of two steps}. The first step is to estimate a North pole. Once the North pole is estimated, a rotation matrix $R$ that can map the estimated North pole to $(0,0,1)$ is left-multiplied to the input image. The left and right images represent the input and output, respectively.} 
    \label{fig:rectification}
\end{figure}

A 360\degree~image covers 180\degree~of vertical field of view (FoV) and 360\degree~of horizontal FoV. 
One of the distinguishing features of a 360\degree~image is to preserve the image information in every direction. 
To exploit this advantageous feature, 360\degree~images are used in popular online platforms such as YouTube and Facebook, which have widely started supporting 360\degree~images or videos~\cite{youtubevr}. 
However, when a 360\degree~ image is captured by an amateur without using any specialized instruments for stabilization (\eg tripod), the 360\degree~image obtained as the output can display slanted objects and wavy horizons due to camera tilts and rolls, as shown in the image on the left of Fig.\ref{fig:rectification}.
If a user views this image using a head-mount display (HMD), he/she is likely to feel falling down or leaning backward. 
This not only diminishes the quality of the virtual reality (VR) experience but can also lead to the user feeling sick. 
The upright adjustment aims to compensate for these tilts and rolls and recover the straight version of the relatively inclined image \cite{lee2013automatic}. 

% -------------------------------Fig1. ------------------------
% -------------------------------Fig1. ------------------------
% -------------------------------Fig2. ------------------------
\begin{figure*}[t]
    \begin{minipage}[b]{1.\linewidth}
        \centering
        \includegraphics[width=1.0\linewidth]{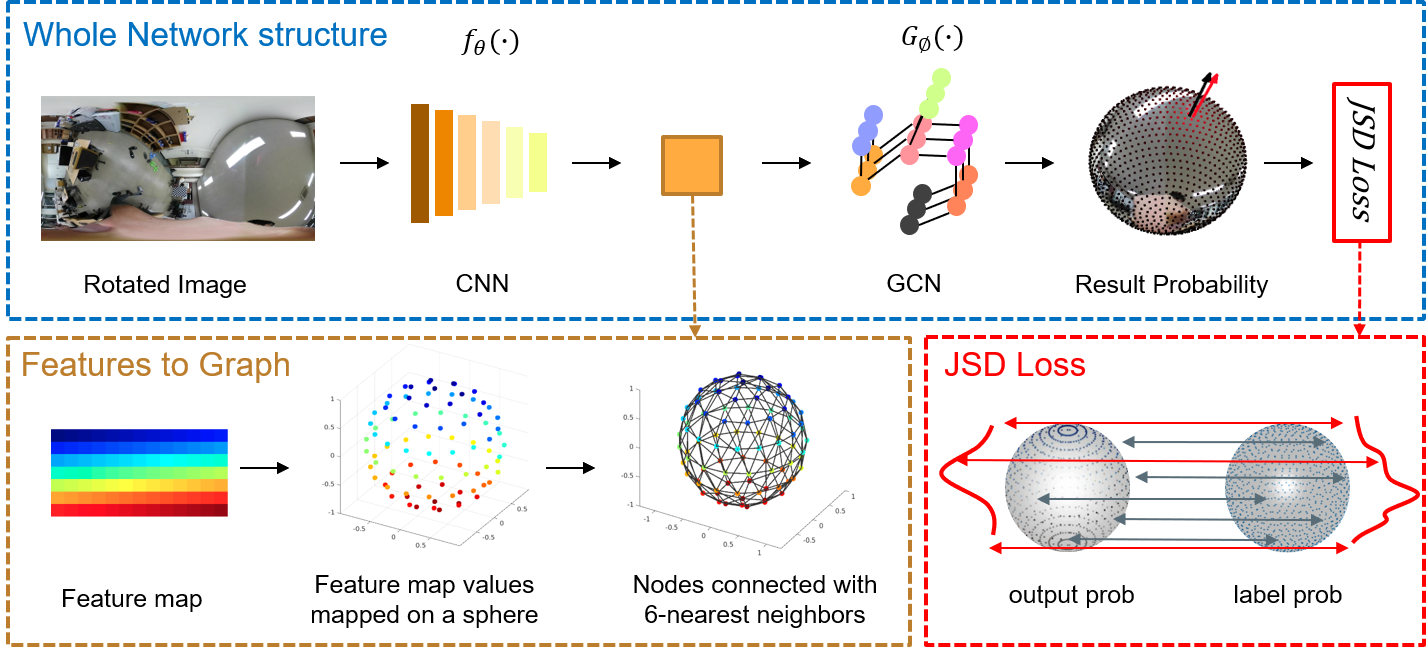}
        \vspace{-7mm}
    \end{minipage}
    \caption{\textbf{Illustration of the network forwarding process}. Orange box explains how the feature map is converted into the graph, wherein the color of the feature map and the nodes of the graph represent the correspondence between the feature map and the points (\ie nodes) on the sphere. After being mapped to the sphere, the nodes are connected to the 6-nearest neighbors to form the graph. Red box illustrates the concept of JSD loss.}
    \label{fig:whole_network}
\end{figure*}
% -------------------------------Fig2. ------------------------
Upright adjustment of 360\degree~images consists of two steps. 
The first step is to estimate a position of a North pole (\ie the opposite direction of gravity). 
The second step is to apply a rotation matrix that can map the estimated North pole to $(0,0,1)$. 
Fig.\ref{fig:rectification} illustrates an example of the upright adjustment. 
Recently, few studies have been conducted on upright adjustment of 360\degree~images based on the deep learning algorithm~\cite{jung2019deep360up,jeon2018deep,shan2019discrete} and have adopted the convolutional neural network (CNN), where the input is a regular grid 2D image. 
As the natural shape of a 360\degree~image is a sphere, each study suggests its own way to fit the 360\degree~image into a regular 2D grid image through projection methods or sampling FoV images. 

In this paper, we investigate a way to process the 360\degree~image in its own natural shape (sphere). 
For processing of spherical data, we adopt the graph convolutional networks (GCN). 
We use the GCN in conjunction with a CNN module. 
The CNN module extracts visual representation from an input image and we convert this feature map into a graph that represents the sphere. 
Finally, the graph is processed by the GCN. 

The main contributions of our method are three-fold. 
\begin{itemize}  
    \item We propose a network composed of the CNN and the GCN. The GCN module helps processing the input image in the form of a sphere.
    \vspace{-1.5mm}
    \item We propose a new loss function. This loss function handles a position of the north pole in a probabilistic way. The loss function reduces a distance between the predicted probability to the ground truth probability of a position of the north pole.
    \vspace{-1.5mm}
    \item We show that our network has reported more competitive result over the typical network composed of the CNN and fully connected layers. The advantages are rotation invariance, fast convergence and the performance.
\end{itemize}

%%%%%%%%%%%%%%%%%%%%%%%%%%%%%%%%%%%%%%%%%%%%%%%%%%%%%%%%%%%%%%%%%%%%%%%%%%%%%%%%%%%%%%%%%%%%%%%%%%%%%%%%%%%%%%%%%
\section{Related Work}
\label{sec:related}

%%%%%%%%%%%%%%%%%%%%%%%%%%%%%%%%%%%%%%%%%%%%%%%%%%%%%%%%%
\subsection{Upright Adjustment Methods}
\noindent\textbf{Feature-based algorithms:}
Feature-based algorithms follow several assumptions to determine features. 
For example, line-based algorithms~\cite{joo2018globally} follow Atlanta world~\cite{schindler2004atlanta} or Manhattan world~\cite{coughlan2001manhattan} assumptions and search for the vanishing point that is most likely in the opposite direction of sky. 
Another kind of feature-based algorithm is based on horizon search\cite{demonceaux2006robust}. 
These algorithms assume that a clearly visible horizon exists in the image and try to find this horizon in the image. 

\noindent\textbf{Deep learning-based algorithms:}
Owing to the ability to extract semantic visual features, CNN-based algorithms are not required to make assumptions in terms of the input. 
However, 360\degree~images have to be fitted into 2D regular grid in order to be processed by the CNN. Existing deep learning papers process flat images generated by projections rather than processing the spherical representation.
Jeon \etal\cite{jeon2018deep} addressed this issue by sampling narrow FoV images from a 360\degree~image. 
Jung \etal\cite{jung2019deep360up} chose the equirectangular projection, which serves to be the most popular choice. 
Yu \etal\cite{shan2019discrete} investigated more accurate projection methods and proposed the discrete spherical image representation.

%%%%%%%%%%%%%%%%%%%%%%%%%%%%%%%%%%%%%%%%%%%%%%%%%%%%%%%%%
\subsection{Graph Convolutional Networks}
GCNs are designed to represent graph structured data such as social networks, 3D meshes, relation database and molecular geometry. 
Most GCNs are trained by propagating information through edges that connect two nodes. 
The connectivity is expressed using the adjacency matrix (\ie square matrix), which represents a finite graph.
Bruna \etal\cite{bruna2013spectral} generalized CNNs into signals defined on graphs. 
Defferrard \etal\cite{defferrard2016convolutional} designed fast localized convolution filters on graphs in the context of spectral graph theory and the Chebyshev polynomial.
Kipf \etal\cite{kipf2016semi} proposed tidy GCNs using first-order approximation of spectral graph convolutions and successfully performed the node classification task. 

%%%%%%%%%%%%%%%%%%%%%%%%%%%%%%%%%%%%%%%%%%%%%%%%%%%%%%%%%%%%%%%%%%%%%%%%%%%%%%%%%%%%%%%%%%%%%%%%%%%%%%%%%%%%%%%%%
\section{Proposed Method}
\label{sec:Proposed}

%%%%%%%%%%%%%%%%%%%%%%%%%%%%%%%%%%%%%%%%%%%%%%%%%%%%%%%%%
\subsection{Method Overview}
The proposed network is composed of a CNN module and GCN module. 
The CNN extracts visual features from the input image in equirectangular projection and the GCN predicts a discrete probability distribution of the North pole, which is represented by a group of points defined on the surface of the sphere sampled by Leopardi \etal\cite{leopardi2006partition}. 
The final predicted position of the North pole is obtained by computing the expectation for $x,y,$ and $z$.

As an input, the 360\degree~image $x  \in {\rm I\!R}^{h\times w\times c}$ is fed into the CNN that produces the feature map $f_{\theta}(x) \in {\rm I\!R}^{h'\times w'\times c'}$. 
Then, the feature map is converted into a graph denoting the spherical representation. 
To convert the feature map into a graph, the map is flattened and projected into the points of the sphere starting from the North pole and moving toward the South pole.
The orange box in Fig.\ref{fig:whole_network} shows the correspondence between the feature map and the graph.

%%%%%%%%%%%%%%%%%%%%%%%%%%%%%%%%%%%%%%%%%%%%%%%%%%%%%%%%%
\subsection{Network Architecture}
For the CNN module, we utilize pre-trained architectures such as ResNet-18~\cite{he2016deep} and DenseNet-121~\cite{huang2017densely}.
The GCN module~\cite{kipf2016semi} is composed of five layers. 
The size of the channel is reduced to half for the consequent layers except the last layer. 
Regardless of the input channel size, the output channel size is $1$ in the last layer. 
In conjunction with the GCN layers, we insert the rectified linear unit (ReLU) in between.
The adjacency matrix is constructed by connecting the $6$-nearest neighbors and is improved into the $n$-hop matrix by multiplying itself for $n$ times. As $n$ grows, it would connect more number of nodes.
Then, the Softmax function is applied to the GCN output, which produces the discrete probability distribution.

\begin{figure}[t]
    \begin{minipage}[b]{1.\linewidth}
        \centering
        \includegraphics[width=1.0\linewidth]{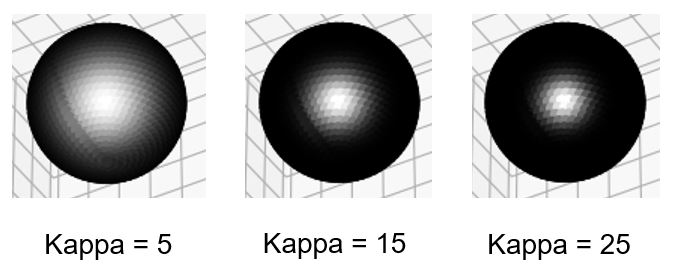}
    \end{minipage}
        \vspace{-3mm}
    \caption{\textbf{Von Mises-Fisher distributions} where $\mu$ is the unit vector heading toward us with different $\kappa$} 
    \label{fig:kappas}
\end{figure}

%%%%%%%%%%%%%%%%%%%%%%%%%%%%%%%%%%%%%%%%%%%%%%%%%%%%%%%%%
\subsection{Objective Function}
We represent the position of the North pole as the probability distribution. 
The output of our networks is a discrete probability distribution of points defined on the surface of the sphere. 
Therefore, it is necessary to generate a probability distribution whose expectation is the ground truth North pole.

%%%%%%%%%%%%%%%%%%%%%%%%%%%%%%%%%%%%%%%%%%%%%%%%%%%%%%%%%
\subsubsection{Distribution Labels}
In directional statistics, von Mises–Fisher distribution is a probability distribution on the $(p-1)$-dimensional sphere. 
The probability density function (PDF) of this distribution for a random $p$-dimensional unit vector is as follows:
\begin{equation}
\label{von_mises_fisher}
    f_{p}(\mathbf{x} ;\kappa\mu^{\mathbf{T}}\mathbf{x}) = C_{p} exp(\kappa \mu^{\mathbf{T}} \mathbf{x}),
\end{equation}
where $\mu$ is the mean direction that is the center of the distribution with $\lVert \mu \rVert= 1$ and $\kappa\geq0$ is a standard deviation of Gaussian distribution on a sphere. 
The larger the value of $\kappa$, the higher is the concentration of the distribution around $\mu$, as shown in Fig.\ref{fig:kappas}. 
Therefore, we set $\mu$ as the ground truth North pole and vary $\kappa$. 
In \eqref{von_mises_fisher}, the normalization constant $C_{p}$($\kappa$) is defined as 
\begin{equation}
\label{von_mises_fisher2}
    C_{p}(\kappa) = \frac{\kappa^{(p/2 - 1)}}{(2\pi)^{p/2} I_{p/2 - 1(\kappa)}}, 
\end{equation}
where $p$ denotes the dimension of the sphere.
By utilizing von Mises-Fisher PDF, we can generate labels for training the network.

%%%%%%%%%%%%%%%%%%%%%%%%%%%%%%%%%%%%%%%%%%%%%%%%%%%%%%%%%
\subsubsection{JSD Loss}
We use the Jenson-Shannon divergence (JSD) as a distance metric and calculate the distance between two probability distributions. 
Then, our method aims to minimize the distance between the predicted distributions and ground truth distributions $label_{dist}$ using the following loss function:
\begin{equation}
\label{jsd_loss}
    \mathcal{L}=JSD(softmax(G_{\phi}(f_{\theta}(x), A)), label_{dist}),
    % JSD(P\parallel Q) = \frac{1}{2} D_{KL}(P\parallel M) + \frac{1}{2} D_{KL}(Q\parallel M)
\end{equation}
where $A$ denotes the adjacency matrix, and $\theta$ and $\phi$ denote the parameters for CNN and GCN, respectively, as shown in Fig.\ref{fig:whole_network}.
The red box in Fig.\ref{fig:whole_network} illustrates the JSD loss concept.

%%%%%%%%%%%%%%%%%%%%%%%%%%%%%%%%%%%%%%%%%%%%%%%%%%%%%%%%%%%%%%%%%%%%%%%%%%%%%%%%%%%%%%%%%%%%%%%
\section{EXPERIMENTS}
\label{sec:experiment}
We used the SUN360 dataset~\cite{xiao2012recognizing}, which consists of 360\degree~images taken in various places (\eg indoor, outdoor, urban, and rural) and different conditions (\eg day and night). 
We sampled $25000$, $5000$, and $4260$ images for training, validation, and testing datasets, respectively. 
All images were synthetically rotated based on the rotation strategy  in~\cite{jung2019deep360up}.

% ------------------------------ Fig4. ------------------------------
\begin{figure*}[t]
    \begin{minipage}[b]{1.\linewidth}
        \centering
        \includegraphics[width=1.0\linewidth]{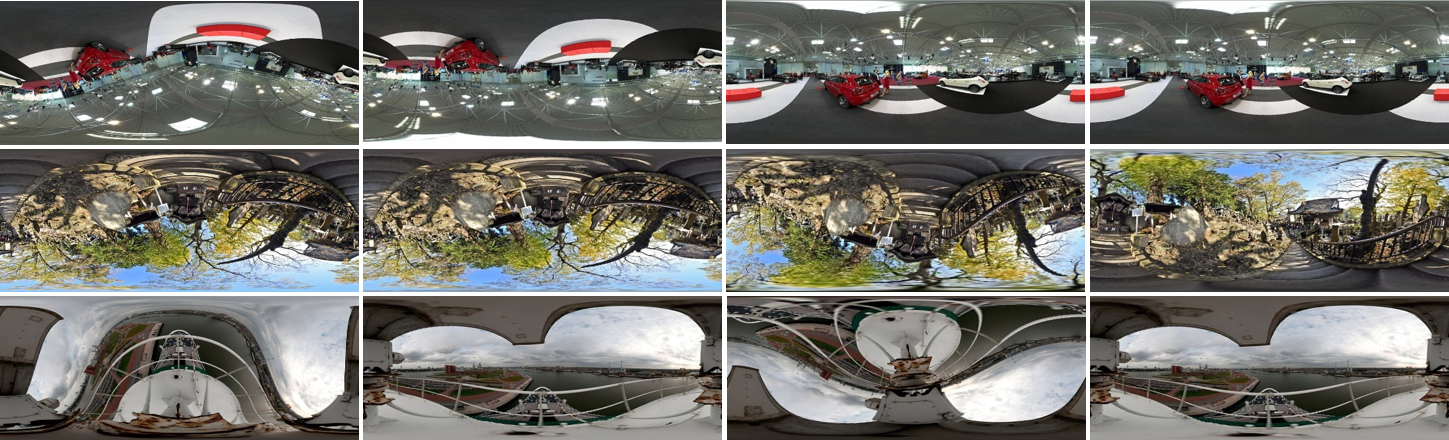}
    \end{minipage}
    \vspace{-3mm}
    \begin{minipage}[b]{0.24\linewidth}
            \centering
            {(a) Input image}
    \end{minipage}
    \begin{minipage}[b]{0.24\linewidth}
            \centering
            {(b) Horizon based \cite{demonceaux2006robust}}
    \end{minipage}
    \begin{minipage}[b]{0.24\linewidth}
            \centering
            {(c) Jung \etal\cite{jung2019deep360up}}
    \end{minipage}
    \begin{minipage}[b]{0.24\linewidth}
            \centering
            {(d) Ours}
    \end{minipage}
    \caption{\textbf{Qualitative Comparison.} 
    In the first row, horizon based method failed because a clearly visible horizon was not detected. 
    In the second row, both horizon based method and Jung\etal failed. 
    In the third row, horizon based method was successful because of a clear horizon, whereas Jung \etal failed.
    In contrast, ours was successful for all the cases and handled various scenarios (\eg nature/urban, indoor/outdoor, and existence of horizon or not).} 
    \label{fig:qualitative_result} 
        \vspace{-3mm}
\end{figure*}

%%%%%%%%%%%%%%%%%%%%%%%%%%%%%%%%%%%%%%%%%%%%%%%%%%%%%%%%%
\subsection{Ablation Study}
 \vspace{-5mm}
% ------------------------------ Table1. ------------------------------
\begin{table}[h]
\caption{We used DenseNet and ResNet as CNN backbones with the different combinations of kappa values. 
DenseNet with kappa value as $25$ shows the best result in terms of the average error. 
The column within $10$\degree indicates the percentage of images whose error is below 10\degree.
}
\begin{center}
\begin{tabular}{c|c|c|c}
\hline
Variants of our method & $\kappa$ & Avg & within 10\degree \\ \hline \hline
DenseNet121 & 15 & 6.0\degree & 90\% \\
DenseNet121 & 20 & 4.3\degree & 97\% \\
DenseNet121 & 25 & \textbf{4.0\degree} & \textbf{97\%} \\
ResNet18 & 15 & 6.4\degree & 93\% \\
ResNet18 & 20 & 6.4\degree & 93\% \\
ResNet18 & 25 & 6.6\degree & 93\% \\ \hline
\end{tabular}
\end{center}
\label{tab:ablation}
\end{table}
 \vspace{-3mm}
We made six variations of our networks by changing their main components, CNN structure, and $\kappa$. 
For the CNN structure, we tested two popular networks, which are ResNet-18~\cite{he2016deep} and DenseNet-121~\cite{huang2017densely}. 
Four different values of $10, 15, 20$, and $25$ were used for the $\kappa$ values, which results in eight combinations. 
According to Table \ref{tab:ablation}, the DenseNet reports better performance than the ResNet. 
This tendency holds with high accuracy regardless of the value of kappa.

\begin{figure}[t]
    \begin{minipage}[b]{1.\linewidth}
        \centering
        \includegraphics[width=0.9\linewidth]{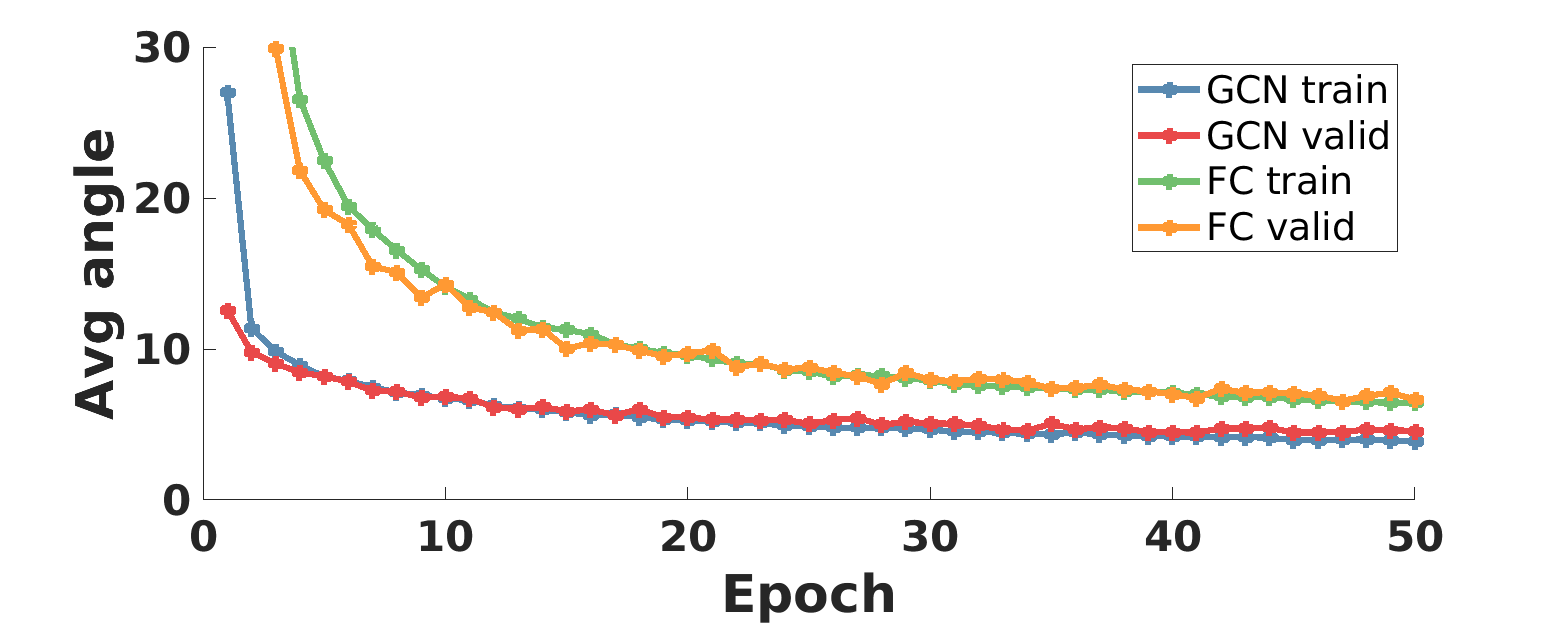}
    \end{minipage}
     \vspace{-7mm}
    \caption{\textbf{Advantages of the GCN module}.} 
    \label{fig:fast_convergence}
\end{figure}

%%%%%%%%%%%%%%%%%%%%%%%%%%%%%%%%%%%%%%%%%%%%%%%%%%%%%%%%%
\subsection{Advantages of the GCN module} 
\vspace{-3mm}
In our method, two primary advantages of using the GCN module are rotation invariance and fast convergence. 
To justify these advantages, we compared our method, which is CNN (DenseNet-121 with kappa value of $25$) $+$ GCN, with CNN $+$ conventional fully-connected layers.

%%%%%%%%%%%%%%%%%%%%%%%%%%%%%%%%%%%%%%%%%%%%%%%%%%%%%%%%%
\noindent\textbf{Rotation Invariance:}
For this experiment, $500$ images were selected from the test set and each image was rotated into $20$ random directions. 
We computed the standard deviation (STD) for each group that shared the same source image. 
A lower standard deviation indicates better rotation-invariant, because, in this case, the error angle remains the same regardless of its initial rotation.
The mean value of STD is \textbf{2.1\degree} for ours and \textbf{4.4}\degree for conventional fully-connected layers-based methods.
The proposed GCN produces more consistent error angles regardless of the initial rotation.  

%%%%%%%%%%%%%%%%%%%%%%%%%%%%%%%%%%%%%%%%%%%%%%%%%%%%%%%%%
\noindent\textbf{Fast convergence:}
Our method with GCN converged much faster than networks with fully connected layers (FC). 
Fig.\ref{fig:fast_convergence} shows the average error of GCN and FC over epochs for the training and validation sets. 
The data has been recorded during a training session. 
Our method with GCN consistently reports better errors for training and validation sets.

%%%%%%%%%%%%%%%%%%%%%%%%%%%%%%%%%%%%%%%%%%%%%%%%%%%%%%%%%
\subsection{Comparison to other methods}
\vspace{-1mm}
\begin{table}[h]
\caption{\textbf{Quantitative Comparison.}  Our network attains the most competitive result according to the average angle. The column within $10$\degree indicates the percentage of images whose error is below 10\degree.}
\begin{center}
\begin{tabular}{@{}c|c|c@{}}
\toprule
 Method & Avg & within $10$\degree \\ \midrule \hline
GCN & \textbf{4.0\degree} & \textbf{97\%} \\
Horizon based \cite{demonceaux2006robust} & 89.7\degree & 20\% \\
Jung \etal\cite{jung2019deep360up} & 5.9\degree & 96\% \\ \bottomrule
\end{tabular}
\label{tab:comparison}
\end{center}
\vspace{-3mm}
\end{table}

We compared our network (\ie DenseNet with $\kappa$ of 25+GCN) with a feature-based algorithm~\cite{demonceaux2006robust} and a deep learning-based algorithm~\cite{jung2019deep360up}. 
We used $4260$ randomly rotated images for testing. 
Table \ref{tab:comparison} demonstrates that our method outperforms other methods in terms of accuracy. 
However, it should be noted that our network is trained for only $50$ epochs, whereas Jung \etal have trained their network for $800$ epochs whose training environments are exactly same with ours.
%. Moreover, the learning rate, optimizer, and training set were exactly same those of Jung \etal\cite{jung2019deep360up}.
\vspace{-3mm}

%%%%%%%%%%%%%%%%%%%%%%%%%%%%%%%%%%%%%%%%%%%%%%%%%%%%%%%%%%%%%%%%%%%%%%%%%%%%%%%%%%%%%%%%%%%%%%%
\section{Conclusion}
\vspace{-3mm}
We present the networks based on the CNN and GCN for upright adjustment. 
The feature map obtained by the CNN is converted into a graph with the spherical representation of the relative input. 
This is the first approach in terms of upright adjustment to preserve its spherical shape. 
With the newly adopted GCN, 
our network shows better rotation invariance and faster convergence over its fully connected layer counterpart.
\label{sec:conclu}

\vspace{-3mm}
\section{acknowledgement} 
\vspace{-3mm}
This work was supported by the Seoul R\&BD Program (CY190032) and (NRF-2018R1A4A1059731).
% This research was supported by the Seoul R\&BD Program(CY190032). 
\label{sec:acknowledgement}

\bibliographystyle{IEEEbib}
\bibliography{refs}

\end{document}